\documentclass[10pt,twocolumn,letterpaper]{article}

\usepackage[pagenumbers]{cvpr} 

\usepackage{graphicx}
\usepackage{amsmath}
\usepackage{amssymb}
\usepackage{pifont}
\usepackage{booktabs}
\usepackage{enumitem}
\usepackage[accsupp]{axessibility}  

\usepackage[pagebackref,breaklinks,colorlinks]{hyperref}

\usepackage[capitalize]{cleveref}
\crefname{section}{Sec.}{Secs.}
\Crefname{section}{Section}{Sections}
\Crefname{table}{Table}{Tables}
\crefname{table}{Tab.}{Tabs.}

\newcommand\blfootnote[1]{%
  \begingroup
  \renewcommand\thefootnote{}\footnote{#1}%
  \addtocounter{footnote}{-1}%
  \endgroup
}

\def\cvprPaperID{18} 
\def\confName{L3D-IVU}
\def\confYear{2023}

\begin{document}

\title{Improving Data-Efficient Fossil Segmentation via Model Editing}


\author{Indu Panigrahi, Ryan Manzuk, Adam Maloof, and Ruth Fong \\ Princeton University, Princeton, NJ, USA \\ {\tt \{indup, rmanzuk, maloof, ruthfong\}@princeton.edu} }

\maketitle

\begin{abstract}
    Most computer vision research focuses on datasets containing thousands of images of commonplace objects. 
    However, many high-impact datasets, such as those in medicine and the geosciences, contain fine-grain objects that require domain-expert knowledge to recognize and are time-consuming to collect and annotate. 
    As a result, these datasets contain few labeled images, and current machine vision models cannot train intensively on them. 
    Originally introduced to correct large-language models, model-editing techniques in machine learning have been shown to improve model performance using only small amounts of data and additional training.
    Using a Mask R-CNN to segment ancient reef fossils in rock sample images, we present a two-part  paradigm to improve fossil segmentation with few labeled images: we first identify model weaknesses using image perturbations and then mitigate those weaknesses using model editing.
    
    Specifically, we apply domain-informed image perturbations to expose the Mask R-CNN’s inability to distinguish between different classes of fossils and its inconsistency in segmenting fossils with different textures. To address these shortcomings, we extend an existing model-editing method for correcting systematic mistakes in image classification to image segmentation with no additional labeled data needed and show its effectiveness in decreasing confusion between different kinds of fossils.
    We also highlight the best settings for model editing in our situation: making a single edit using all relevant pixels in one image (vs. using multiple images, multiple edits, or fewer pixels).
    Though we focus on fossil segmentation, our approach may be useful in other similar fine-grain segmentation problems where data is limited.
\end{abstract}

\setlength{\textfloatsep}{10pt}
\setlength{\floatsep}{10pt}

\section{Introduction}
\label{sec:intro}
\begin{figure}[hbt]
\centering
\includegraphics[width=\linewidth]{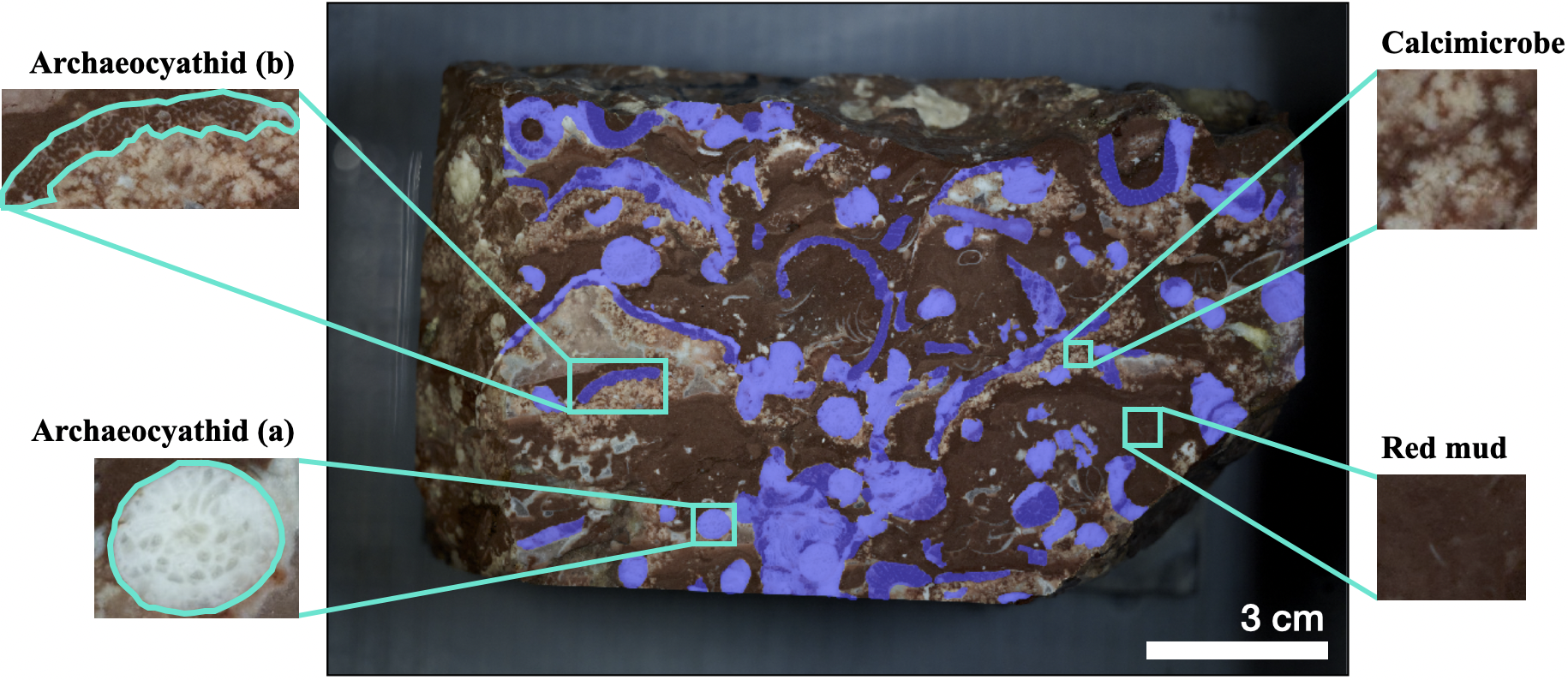}
\caption{\textbf{Overview of rock sample.} Sample annotated image from our rock sample dataset with magnified examples of archaeocyathids, a patch of calcimicrobe, and red mud. There also exist other classes that are omitted from this paper for simplicity. We denote two primary textures of archaeocyathids by \textbf{(a)} and \textbf{(b)}; \textbf{(a)} has a discernible porous texture, while \textbf{(b)} is filled with red mud and consequently blends in with the surrounding mud. Archaeocyathid pixels in the full image are colored in purple with 80\% transparency; magnified examples show the original coloring.}
\label{fig:fewfossils}
\end{figure}

Today, most computer vision models are trained on large-scale datasets (e.g.\ ImageNet~\cite{imagenet} and Microsoft COCO~\cite{cocodataset}) that contain thousands of annotated images of commonplace scenes and objects. 
This trend exists in part because deep neural networks, the current state-of-the-art in machine learning, require large amounts of data in order to learn complex, highly predictive patterns that enable them to outperform classical machine learning methods.
However, many high-impact domains, such as those in the natural and life sciences, involve fine-grain objects that require domain-expert knowledge to recognize and are time-consuming to collect and annotate~\cite{botany,geosciences,medicine}. 
As a result, these datasets contain few labeled images. However, deep neural networks often cannot be sufficiently trained on small datasets to segment objects in images well (\cref{tab:my_label}).

Model-editing methods have recently emerged in natural language processing (NLP)~\cite{meng2022locating, zhu2020,meng2022memit} and now computer vision~\cite{sinitsin2020,ilharco2022,editingclassifier} as a way to correct for systematic mistakes in deep learning models.
These techniques differ from related work in domain adaptation and continual learning because of their focus on correcting mistakes in models (as opposed to adapting to a domain shift in inputs and/or retaining memory of the original distribution)
by using limited data and additional training.
This combination of limited data and focus on correcting errors makes model editing particularly promising in domains like geosciences, where data is hard to come by, yet a small amount of domain expertise can be accessed to correct and improve a model.

\begin{figure}[hbt]
\centering
\includegraphics[width=0.8\linewidth]{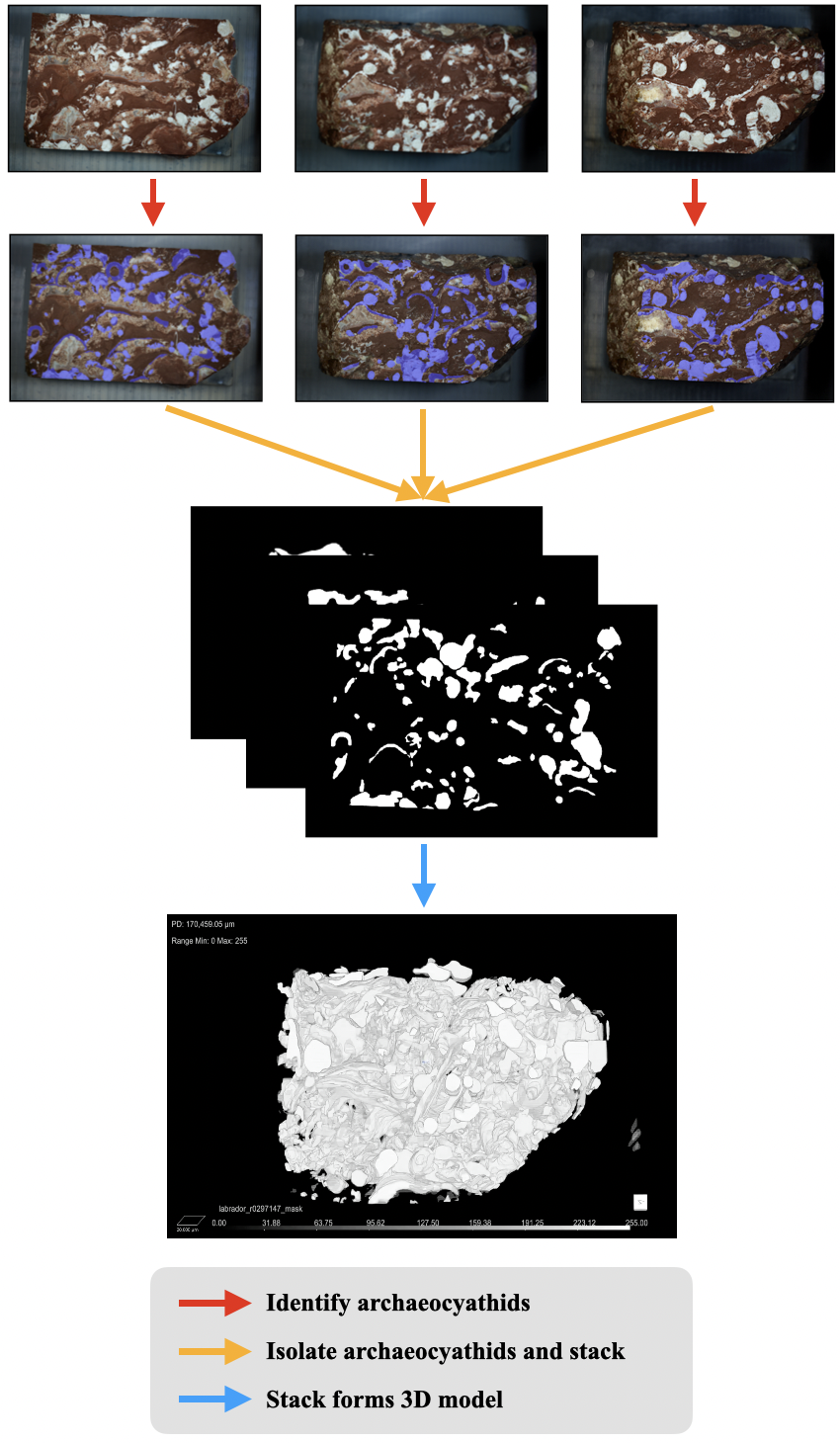}
\caption{\textbf{Fossil modeling process.} From top to bottom: The archaeocyathids in each image in the stack are segmented, and the segmented portions are stacked to form a 3D model.}
\label{fig:reconstruction}
\end{figure}

In this paper, we present a two-part framework for improving a fossil segmentation model. First, we use domain-informed image perturbations to \textbf{identify model weaknesses}. Second, we adapt a model-editing method to \textbf{mitigate those weaknesses}. We focus our work on improving a Mask R-CNN~\cite{maskrcnn} trained on a small set of annotated rock sample images to segment ancient reef fossils (\cref{fig:fewfossils}).
We are interested in segmenting \textit{archaeocyathids} (\cref{fig:fewfossils}), an extinct reef-building sponge~\cite{archaeos}. 
Studying these ancient reef fossils would allow us to understand their influence on past oceanic biodiversity~\cite{manzuk} and inform our understanding of the impact that dwindling coral reefs today will have on Earth's future climate and biosphere~\cite{coralbleaching}.
In many cases, as with our rock sample, embedded specimens are too delicate to be physically isolated from the surrounding material. 
One solution is to generate 3D models of the specimens from serial sectioning and imaging of samples~\cite{cloudina}.
To build such models, we need to segment the pixels of archaeocyathids in each image and then stack the resulting masks (\cref{fig:reconstruction}).

Due to the fine-grain appearance of archaeocyathids and the domain knowledge needed to recognize them, manually segmenting each image is time-consuming, with one image taking around 2 hours to annotate (\cref{tab:my_label}). 
Given that a single, full image stack contains over 3,000 images, manually segmenting all images is labor intensive.
Furthermore, there exist many rock samples archived in museums and universities, and the image stacks generated for each can vary significantly in terms of visual appearance.
Thus, we seek to automate the segmentation process by utilizing the Mask R-CNN model~\cite{maskrcnn} to segment an image stack from a limited amount of labeled data ($\approx 10 $ images). Because of this data constraint, simply fine-tuning our model did not produce high-quality segmentation masks that are needed when forming 3D models for such specimens.

\begin{table}[hbt]
    \centering
    \begin{tabular}{l|c|c}
    \toprule
    & MS COCO & Ours \\
    \midrule
         Number of labeled images & $>200$K & $10$ \\
         Time to annotate an image (hrs) & $0.66$~\cite{googleblog} & $2$ \\
         Domain expert knowledge needed & \ding{55} & \ding{51} \\
    \bottomrule
    \end{tabular}
    \caption{Comparison between COCO dataset~\cite{cocodataset} and ours.}
    \label{tab:my_label}
\end{table}

Rather than annotating more images, we focused on leveraging a model-editing technique~\cite{editingclassifier} to improve our baseline model.
This technique requires no additional labelled data, making it particularly well-suited for specialized datasets like ours that require domain knowledge expertise and significant time to annotate.

In this work, we present a two-part, data-efficient paradigm that combines domain-informed image perturbations with a model-editing method to first \textbf{identify} and then \textbf{mitigate weaknesses} in our model.
Our main contributions are summarized as follows:
\begin{itemize}[nolistsep]
    \item We first identify model weaknesses via image perturbations and texture synthesis.
    From these experiments, we found two main weaknesses. 
    First, our model often confuses archaeocyathids with other visually-similar types of fossils (e.g. \textbf{interclass confusion}).
    Second, our model is not robust to the visual diversity that archaeocyathids can have (e.g. \textbf{intraclass variation}).
    \item We then mitigate the identified model weaknesses by extending and evaluating an existing, model-editing technique~\cite{editingclassifier} for correcting systematic mistakes in image classification to image segmentation. 
    In particular, we find that certain edits improve the model's ability to distinguish between archaeocyathids and other types of fossils.
    \item Lastly, we gain several insights on how to effectively use the editing method.
    We show that performing a single edit using one image (vs. using multiple images or multiple, sequential edits) is sufficient and, further, that editing by using all relevant pixels (vs. a smaller subset of pixels) yields the best results.
\end{itemize}

\section{Related Work}
\label{sec:related}
In this section, we first discuss related work in the field of \emph{connectomics}, which tackles a similar problem of building a 3D structure from 2D data.
Then, we discuss relevant literature for the two core parts of our work: identifying model weaknesses via image perturbations (e.g. \emph{image occlusion} and \emph{texture synthesis}) and  mitigating weaknesses via \emph{model editing}.

\noindent\textbf{Connectomics.} 
Connectomics tackles a similar problem to ours when learning the 3D structure of neurons from 2D brain scan images.
The reconstruction process involves delineating boundaries around regions in the scans just as we segment archaeocyathids~\cite{connectomicsoverview}.
However, most work in connectomics has been directed towards creating novel network architectures~\cite{connectomicsex1,connectomicsex2,connectomicsex3} rather than using interpretability or model-editing techniques to understand and mitigate model failures. 
Similar to approaches in connectomics, we modified our Mask R-CNN to leverage similarities between neighboring images in the stack; however, we found that our model performed poorly despite this modification.

\noindent\textbf{Image occlusion.} 
Image occlusion involves occluding part of an input image and observing the resulting effect on a model's output decision.
Several works utilize image occlusions to generate attribution heatmaps that visualize the most important image regions for a model's decision~\cite{zeiler2014visualizing,ribeiro2016should,occludeuntilmisclassify,fong17interpretable,Petsiuk2018rise}.
Others partially occlude images during training as a data augmentation technique to improve model robustness~\cite{hideandseek,devries2017improved,fong2019occlusions} and/or localization performance~\cite{wei2017object}.
Our work is more similar to those using occlusions to generate attribution heatmaps, as we selectively occlude all pixels from certain classes and observe the effect on the model to identify its shortcomings. 
However, we further use the perturbed images to edit our model.

\noindent\textbf{Texture synthesis.}
Texture synthesis refers to methods that generate a synthetic, often realistic-looking texture~\cite{gatys2015texture,texturesynthesis}.
It can be used in a variety of ways: from inpainting a corrupted image~\cite{deepimageprior}, to visualizing what kind of visual features most activates a channel in a network (i.e. feature visualization)~\cite{mordvintsev2018differentiable}, to studying a network's relative bias towards texture vs.\ shape~\cite{geirhos2018imagenettrained}.
More similar to feature visualization, we generate certain textures in order to study how our model responds to the visual appearance of archaeocyathids.

\begin{figure}[t!]
\centering
\includegraphics[width=\linewidth]{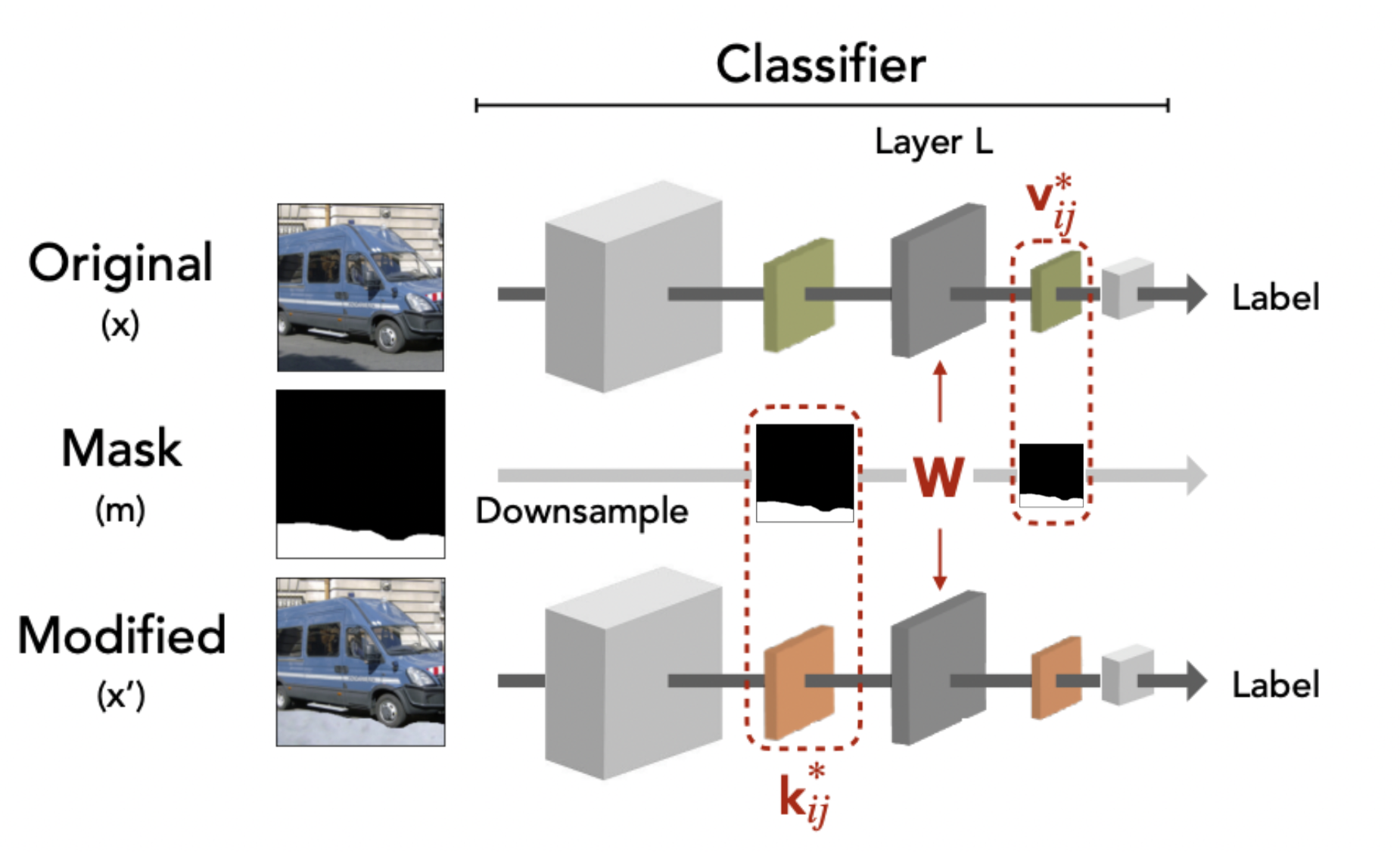}
\caption{\textbf{Overview of model-editing method~\cite{editingclassifier}.} Layer $L$ is the targeted convolutional layer, $k^*$ is the input representation (keys) of the snow-covered road, and $v^*$ is the output activation (values) of the paved road. The method maps $k^*$ to $v^*$ by editing the weights $w$ of the convolutional layer $L$. A mask $m$ can be used to restrict the edit to affect the road pixels only; we do not use a mask for any of our edits. Figure adapted from~\cite{editingclassifier}.}
\label{fig:editingclassifier}
\end{figure}

\noindent\textbf{Model editing.}
There have been a number of methods proposed for editing a model after a pretraining period in computer vision and NLP.
From the machine learning fairness literature, several works have proposed to debias a model so that sensitive demographic information (e.g. race and gender) does not inform model predictions~\cite{beutelfair,zhang2018mitigating,adv_fairness,laftr,wang2018balanced}.
However, not all model errors relate to a societal bias.

In addition to correcting for model biases, model-editing techniques have been used to update the knowledge encoded in large language models to remove outdated information and/or to introduce new information \cite{meng2022locating, zhu2020,meng2022memit}.
Similarly, in computer vision, model-editing techniques have been introduced for correcting mistakes in image classification \cite{sinitsin2020,ilharco2022,editingclassifier}.
However, little work has been done towards editing image segmentation models.

One recent work by Santurkar et al.~\cite{editingclassifier} proposes editing an object classifier to correct for systematic mistakes, like misclassifying vehicles on snow.
In this example, they map a synthesized snow texture underneath vehicles to a more typical asphalt road pattern such that the edited model classifies vehicles on snow as accurately as it classifies vehicles on asphalt.
Their method can edit a model using a single image and a corresponding perturbed version, so it can be adapted to models trained on small datasets.

Specifically, for a selected convolutional layer, they refer to its input as \textit{keys} and its output as \textit{values} (\cref{fig:editingclassifier}).
Then, they use an L1 loss function to edit the weights of the layer such that the keys for the image with the snow-covered road map to the values for the image with the paved road after the snow-covered road passes through the layer. 
Using a rock sample dataset, we extend this method to image segmentation by first applying domain-informed image perturbations to identify systematic mistakes that the Mask R-CNN makes and then editing the model to correct those mistakes.

\section{Experimental Setup}
\label{sec:setup}
\noindent\textbf{Dataset.} 
We use images of an archaeocyathid-bearing rock sample from \cite{manzuk} which were shared by their authors.
The dataset was collected by alternately grinding and imaging cross sections of the sample \cite{imagingrocks}. 
Each image depicts a cross section of the rock sample and contains pixels that represent red mud and the embedded remains of different types of fossils (\cref{fig:fewfossils}). 
In this paper, we focus on archaeocyathid and calcimicrobe fossils  (Fig. \ref{fig:fewfossils}).

We annotate a total of 10 images by tracing an individual instance (polygon) for each archaeocyathid\cite{cocoannotator}. 
We split our 10 annotated images into the following subsets: 6 training, 2 validation, 2 test.

\noindent\textbf{Model.}
We fine-tune a Mask R-CNN pretrained on ImageNet and COCO~\cite{detectron2} on our 6 training images to perform instance segmentation for individual archaeocyathids.
Similar to approaches in connectomics~\cite{connectomicsex2}, we modified our model to leverage the fact that the archaeocyathids remain in similar locations between close layers in the stack by influencing the ranking of the proposal boxes generated by the Region Proposal Network in the Mask R-CNN.
However, we found that the model still classified several non-archaeocyathid fossils as archaeocyathids and generally did not produce precise masks for archaeocyathids.

The precisions of the archaeocyathid masks are particularly important because the identification of spurious non-archaeocyathid pixels interferes with measurements on the rendered 3D model and consequently provides misleading information about the reef's structure.
We would prefer that a few archaeocyathids are missed rather than being fully segmented in an instance containing non-archaeocyathid pixels. 
In other words, we prioritize precision over recall.
Thus, in this work, we focus on leveraging model-editing, with a specific goal of improving the precision of segmentation masks.




\section{Addressing Interclass Confusion}
\label{sec:interclass}
\subsection{Identify model weakness}
\begin{figure}[hbt]
\centering
\includegraphics[width=0.8\linewidth]{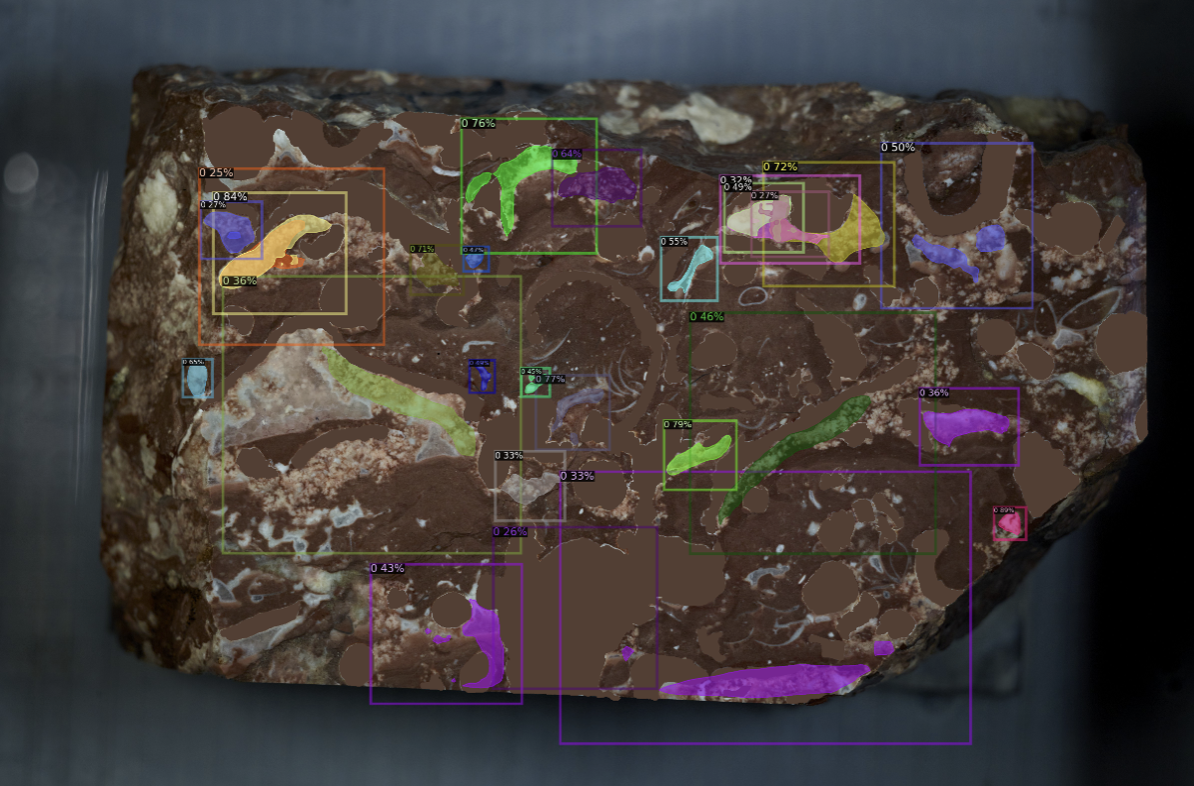}
\caption{\textbf{Example of interclass confusion.} Resulting segmentation when archaeocyathids are inpainted with a shade of red mud. The Mask R-CNN misclassifies several instances of calcimicrobe (boxed and filled with various colors) as archaeocyathids.}
\label{fig:occludearchaeos}
\end{figure}

\paragraph{Archaeocyathid vs.\ non-archaeocyathid fossil confusion.}
For our dataset, the Mask R-CNN sometimes labels instances of another fossil called \textit{calcimicrobe} (Fig.~\ref{fig:fewfossils}) along with a few other non-archaeocyathid fossils as archaeocyathids. 
To analyze this trend, we occlude all archaeocyathids from an image by inpainting them with a shade of red mud that we extract from a manually-selected red mud pixel. 
While the Mask R-CNN ideally should identify no archaeocyathids in the perturbed image, it instead classifies large portions of calcimicrobe as archaeocyathids (Fig.~\ref{fig:occludearchaeos}). 
Thus, the Mask R-CNN cannot clearly distinguish between archaeocyathids and calcimicrobes.

\begin{figure}[hbt]
\centering
\includegraphics[width=0.7\linewidth]{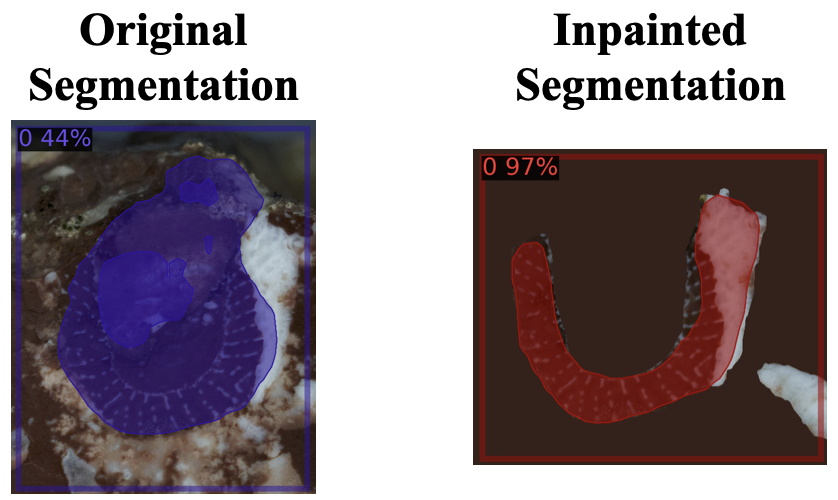}
\caption{\textbf{Example of improved masks from inpainted image.} The mask of this archaeocyathid improves when the original image (left) is inpainted with red mud (right).}
\label{fig:inpaintimprovementex}
\end{figure}

\vspace{-1em}
\paragraph{Archaeocyathid vs.\ red mud separability.}
As a complementary occlusion, we inpaint all non-archaeocyathid pixels with a shade of red mud (Fig.~\ref{fig:inpaintall}) in our 6 training images and run inference. 
The quality of the instance masks drastically improves (mean instance-level IoU across all 443 archaeocyathids from training images increases from $0.63 \pm 0.29$ to $0.78 \pm 0.24$, mean instance-level precision increases from $0.78 \pm 0.22$ to $0.89 \pm 0.17$, mean instance-level recall increases from $0.78 \pm 0.25$ to $0.86 \pm 0.20$) (Fig.~\ref{fig:inpaintimprovementex}).
Thus, the model generally can distinguish between archaeocyathids and a simplified version of red mud.

Since we only need to isolate the archaeocyathid pixels, we have a binary segmentation task with archaeocyathids as positive pixels and non-archaeocyathids as negative ones.
Thus, it would be ideal if the model associated all negative pixels with a concept it already recognizes, namely red mud.

\subsection{Mitigate model weakness}
\begin{figure}[hbt]
\centering
\includegraphics[width=0.9\linewidth]{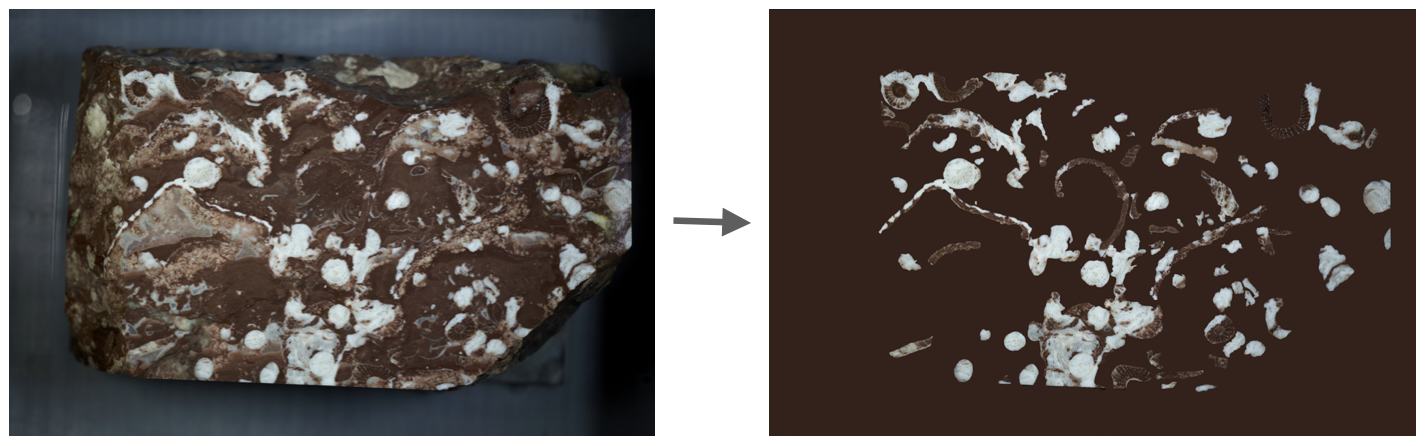}
\caption{\textbf{Mapping non-archaeocyathids to red mud.} We apply the model-editing method to map the input representation of the unperturbed image (left) to the output representation of the inpainted image (right) to enforce a binary supercategorization of archaeocyathids and non-archaeocyathids.}
\label{fig:editingclassifierinpaint}
\end{figure}

\begin{figure}
\centering
     \begin{subfigure}[b]{0.9\linewidth}
         \includegraphics[width=\linewidth]{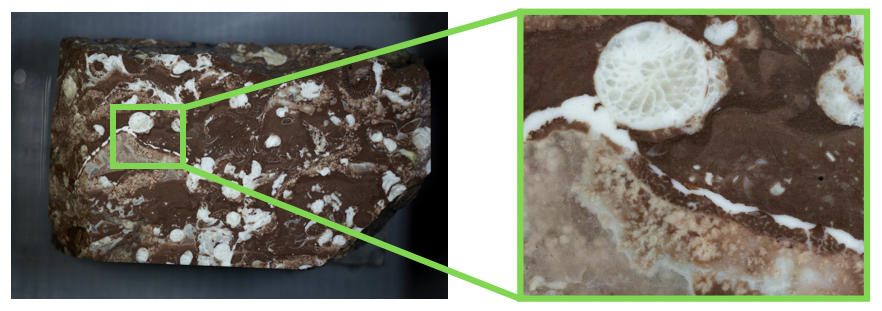}
         \caption{Original image.}
         \vspace{1em}
         \label{fig:original}
     \end{subfigure}
     \begin{subfigure}[b]{0.9\linewidth}
         \includegraphics[width=\linewidth]{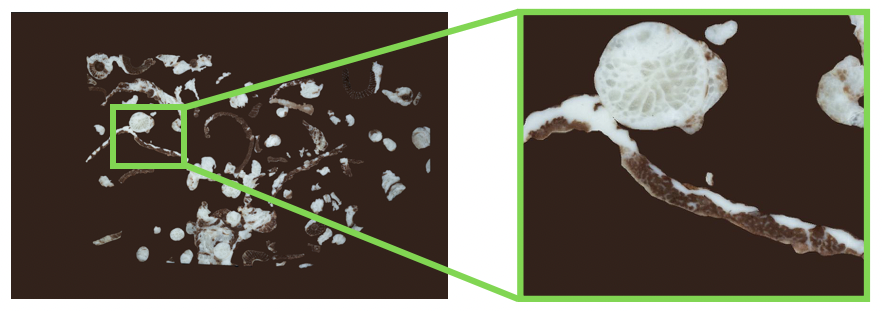}
         \caption{Inpaint all non-archaeocyathids with a shade of red mud.}
         \vspace{1em}
         \label{fig:inpaintall}
     \end{subfigure}
     \begin{subfigure}[b]{0.9\linewidth}
         \includegraphics[width=\linewidth]{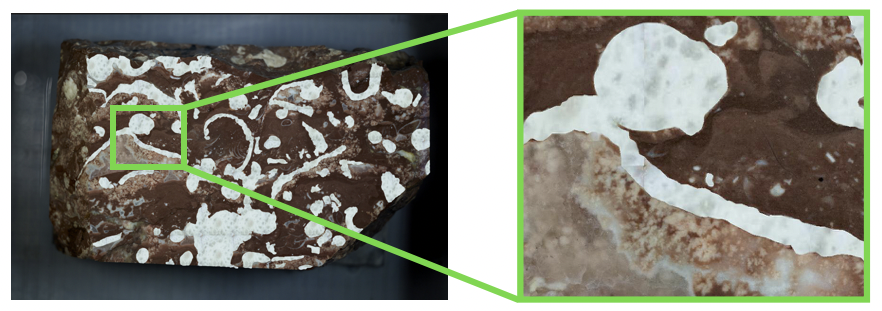}
         \caption{Substitute porous texture for all archaeocyathids.}
         \vspace{1em}
         \label{fig:goodtextures}
     \end{subfigure}
     \begin{subfigure}[b]{0.9\linewidth}
         \includegraphics[width=\linewidth]{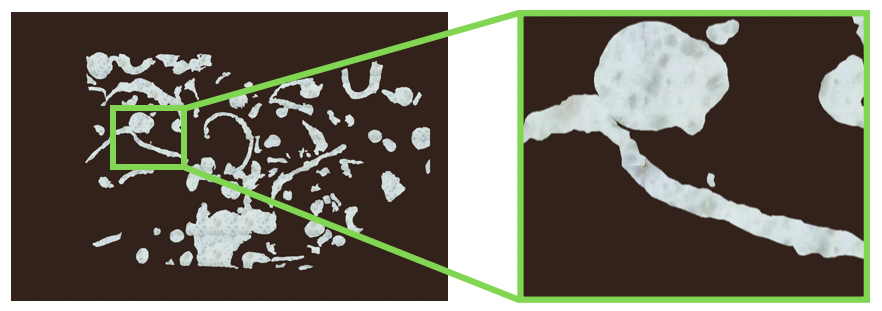}
         \caption{Inpaint (b) and (c) simultaneously.}
         \vspace{1em}
         \label{fig:inpaintsimultaneous}
     \end{subfigure}
     \caption{\textbf{Image perturbations.} (a) shows the original training image with magnified archaeocyathids, (b) shows the version of the image with all non-archaeocyathid pixels inpainted with a manually-extracted solid shade of red mud, (c) shows the version with only the archaeocyathids replaced with a porous texture, and (d) shows the version with (b) and (c) simultaneously.}
\end{figure}

\paragraph{Mapping non-archaeocyathids to red mud to reduce interclass confusion.}
To enforce this binary supercategorization, we apply the model-editing method~\cite{editingclassifier} to one training image such that the model is encouraged to associate all non-archaeocyathid pixels with red mud (Fig.~\ref{fig:editingclassifierinpaint}). 
Specifically, our $k^*$ (Fig.~\ref{fig:editingclassifier}) is the input representation of the original image (Fig.~\ref{fig:original}), and our $v^*$ is the output representation of the same image with all non-archaeocyathid pixels inpainted with red mud (Fig.~\ref{fig:inpaintall}).
We perform $20k$ rewriting steps at a learning rate of $10^{-4}$.
Furthermore, we try editing with each of the 6 training images individually.

\vspace{-1em}
\vspace{-1em}

The model-editing method applies to feature maps, so we edit the weights of \emph{each} of the 5, 3x3 output convolutional layers in the ResNet-101 FPN backbone~\cite{fpn} that produce the feature maps for the Mask R-CNN. 
Doing so means that we perform the edits at different resolutions and can consequently target objects of various sizes in the image. 

\begin{figure}[t!]
\centering
\includegraphics[width=0.4\linewidth]{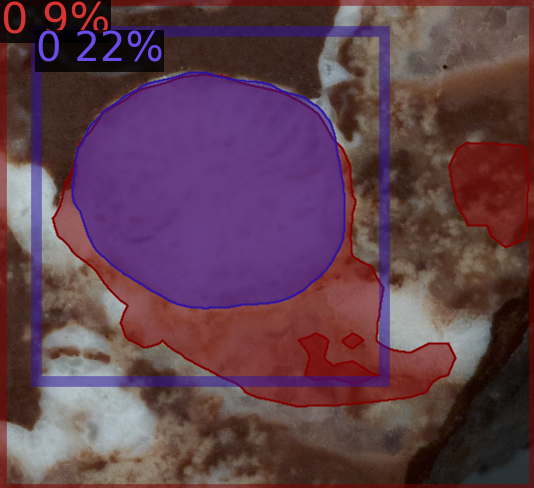}
\caption{\textbf{Example of overlapping instance masks.} The original model predicts two instance masks (one colored in purple and another colored in red) for this archaeocyathid.}
\label{fig:overlappingmasks}
\end{figure}

\paragraph{Evaluation details.} 
Since the validation set was not used when applying the model-editing method, we combine our validation and test sets to evaluate the mapping on more archaeocyathids.
Specifically, we obtain the mean instance-level precision, recall, and IoU across all 215 archaeocyathids in the 4 images using a confidence threshold of 0.
We use this threshold because we wish to evaluate whether or not the Mask R-CNN classifies a pixel as an archaeocyathid at all.
Since an archaeocyathid sometimes has several, overlapping predicted masks (Fig.~\ref{fig:overlappingmasks}), we match each ground truth instance to the predicted instance mask with the highest IoU and take the mean across the matched predicted masks (i.e. the identified archaeocyathids) rather than across all predicted masks to avoid inflating our results.

\vspace{-1em}
\paragraph{Precision of archaeocyathid masks improves.}
We find that the precisions of archaeocyathid masks improve significantly, while the IoU and recall scores decrease (\cref{tab:nonarchresults}).
We also separately evaluate the mapping on each edit image and observe a similar trend except that the mean IoU improves for some images.
For our application, precision is the most important metric because the false identification of extraneous fossils as archaeocyathids interferes with measurements on the rendered 3D model.
Thus, the mapping does reduce interclass confusion, even if it decreases the coverage of archaeocyathid pixels.

The choice of training image does seem to impact the performance; for example, editing with $C$ improves the precision more than editing with $E$ (\cref{tab:nonarchresults}).
Additionally, we try mapping different amounts of non-archaeocyathid pixels to red mud, using image $C$ because it produces the best edited model.
We find that mapping more non-archaeocyathid pixels to red mud produces more precise masks (\cref{tab:onenonarchresults}).
This trend suggests that mapping all the non-archaeocyathid pixels at once is more effective than mapping a small portion.

\begin{table}
  \centering
  \begin{tabular}{c|c|c|c}
    \toprule
    Image & Precision & Recall & IoU\\
    \midrule
    None &
    $0.86 \pm 0.17$ &
    $\mathbf{0.75 \pm 0.26}$ & 
    $\mathbf{0.63 \pm 0.28}$\\
    
    \midrule
    
    A & 
    $\mathbf{0.90 \pm 0.13}$ &
    $0.59 \pm 0.26$ & 
    $0.52 \pm 0.27$ \\
    
    B & 
    $\mathbf{0.88 \pm 0.17}$ &
    $0.59 \pm 0.25$ &
    $0.52 \pm 0.27$ \\
    
    C & 
    $\mathbf{0.91 \pm 0.12}$ &
    $0.63 \pm 0.26$ &
    $0.56 \pm 0.27$ \\
    
    D & 
    $\mathbf{0.89 \pm 0.15}$ &
    $0.64 \pm 0.25$ &
    $0.56 \pm 0.27$ \\
    
    E & 
    $\mathbf{0.87 \pm 0.17}$ &
    $0.64 \pm 0.26$ &
    $0.56 \pm 0.28$ \\
    
    F & 
    $\mathbf{0.90 \pm 0.15}$ &
    $0.65 \pm 0.26$ &
    $0.57 \pm 0.27$ \\
    
    \bottomrule
  \end{tabular}
  \caption{\textbf{Mapping non-archaeocyathids to red mud.} 
  Metrics computed on 215 archaeocyathids from 4 images (mean and standard deviation reported) when editing with each of the training images. 
  The ``Image" column denotes the training image that was used for editing. 
  The top row indicates the original model's performance on the test images (no edits).
  Precision improves when editing with any training image, while recall and IoU decrease.
  }
  \label{tab:nonarchresults}
\end{table}

\subsection{Editing with Multiple Images}
\label{sec:nonarchextension}
In addition to editing with different images individually, we test the effect of editing with more than one image.
\vspace{-1em}
\paragraph{Experimental details.}
We experiment with five additional mappings, each of which incorporates a new edit image.
We again use a learning rate of $10^{-4}$ and perform $20k$ rewriting steps for each image.
For example, $A,B$ edits with image $A$ for $20k$ steps with $lr = 10^{-4}$ followed by image $B$ for an additional $20k$ steps at the same learning rate.
We add images in order of increasing percentage of archaeocyathid pixels, so image $A$ contains the lowest percent of archaeocyathid pixels, and image $E$ contains the highest percent of archaeocyathid pixels.
Furthermore, we test sequences in increasing and decreasing order of precision, recall, and IoU (without incrementally adding images).

\vspace{-1em}
\paragraph{Editing with one image is sufficient.}
The performance of the model edited with a combination loosely corresponds to the performance of the model edited with the last image in the combination.
For example, the performance of the model edited with $A,B,C$ is identical to that of the model edited just with $C$ (\cref{tab:nonarchresults,tab:extendednonarchresults}).
More generally, the mean instance-level metrics are similar to those under the model edited with the last image in the combination.
One exception is the model edited with $A,B$ which performs worse overall.
Thus, we find that editing the model with one inpainted image is sufficient.

\setlength{\tabcolsep}{5pt}
\begin{table}
  \centering
  \begin{tabular}{c|c|c|c}
    \toprule
    \% Pixels & Precision & Recall & IoU\\
    \midrule
    None &
    $0.86 \pm 0.17$ &
    $\mathbf{0.75 \pm 0.26}$ & 
    $\mathbf{0.63 \pm 0.28}$\\
    
    \midrule
    1 &
    $0.86 \pm 0.16$ &
    $0.75 \pm 0.26$ & 
    $0.63 \pm 0.28$\\
    
    35 & 
    $0.90 \pm 0.14$ &
    $0.65 \pm 0.29$ &
    $0.57 \pm 0.31$ \\
    
    100 & 
    $\mathbf{0.91 \pm 0.12}$ &
    $0.63 \pm 0.26$ &
    $0.56 \pm 0.27$ \\
    
    \bottomrule
  \end{tabular}
  \caption{\textbf{Mapping different amounts of non-archaeocyathid pixels to red mud.} 
   Metrics computed on 215 archaeocyathids from 4 images (mean and standard deviation reported).
  ``\% Pixels" indicates the percent of non-archaeocyathid pixels in image $C$ that were inpainted with red mud. 
  The first row shows performance of the original model (no edits). 
  The second row is when pixels for one calcimicrobe are replaced.
  The third row is when non-archaeocyathid pixels on the rock face (i.e. excluding the sides of the rock and the platform on which the rock sits) are replaced.
  The last row is when all non-archaeocyathid pixels are replaced.
  When more non-archaeocyathid pixels are replaced, the precision of the archaeocyathid masks improves.
}
  \label{tab:onenonarchresults}
\end{table}
\setlength{\tabcolsep}{2pt}
\begin{table}
  \centering
  \begin{tabular}{l|c|c|c}
    \toprule
    Sequence & Precision & Recall & IoU\\
    \midrule
    None &
    $0.86 \pm 0.17$ &
    $\mathbf{0.75 \pm 0.26}$ & 
    $\mathbf{0.63 \pm 0.28}$\\
    
    \midrule
    
    A,B & 
    $0.85 \pm 0.24$ &
    $0.39 \pm 0.25$ &
    $0.35 \pm 0.24$ \\
    
    A,B,C & 
    $\mathbf{0.91 \pm 0.12}$ &
    $0.63 \pm 0.26$ &
    $0.56 \pm 0.27$ \\
    
    A,B,C,D & 
    $\mathbf{0.90 \pm 0.14}$ &
    $0.64 \pm 0.25$ &
    $0.56 \pm 0.27$ \\
    
    A,B,C,D,F & 
    $\mathbf{0.89 \pm 0.16}$ &
    $0.64 \pm 0.26$ &
    $0.56 \pm 0.28$ \\
    
    A,B,C,D,F,E & 
    $\mathbf{0.87 \pm 0.18}$ &
    $0.67 \pm 0.25$ &
    $0.56 \pm 0.28$ \\
    
    \midrule
    
    E,B,D,A,F,C & 
    $\mathbf{0.91 \pm 0.13}$ &
    $0.62 \pm 0.27$ &
    $0.55 \pm 0.28$ \\
    
    C,F,A,D,B,E & 
    $\mathbf{0.88 \pm 0.15}$ &
    $0.65 \pm 0.26$ &
    $0.57 \pm 0.27$ \\
    
    \midrule
    
    A,B,C,E,D,F & 
    $\mathbf{0.90 \pm 0.14}$ &
    $0.63 \pm 0.26$ &
    $0.57 \pm 0.27$ \\
    
    F,D,E,C,B,A & 
    $\mathbf{0.89 \pm 0.15}$ &
    $0.59 \pm 0.26$ &
    $0.51 \pm 0.27$ \\
    
    \midrule
    
    B,A,E,D,C,F & 
    $\mathbf{0.89 \pm 0.15}$ &
    $0.64 \pm 0.25$ &
    $0.57 \pm 0.27$ \\
    
    F,C,D,E,A,B & 
    $\mathbf{0.87 \pm 0.16}$ &
    $0.62 \pm 0.25$ &
    $0.52 \pm 0.27$ \\
    
    \bottomrule
  \end{tabular}
  \caption{\textbf{Mapping non-archaeocyathids to red mud using multiple, sequential edit images.}
  Metrics computed on 215 archaeocyathids from 4 images (mean and standard deviation reported).
  The ``Sequence" column denotes the sequence of training images used for each edit. 
  The first set of 6 sequences corresponds to adding one image at a time.
  The next set of 2 sequences is in order of increasing and decreasing precision when using a single image (\cref{tab:nonarchresults}).
  The next two sets are in order of increasing and decreasing recall and IoU respectively.
  Most combinations are comparable to editing with the last image only (\cref{tab:nonarchresults}) and generally improve precision while decreasing recall and IoU.
  }
  \label{tab:extendednonarchresults}
\end{table}

\section{Addressing Intraclass Inconsistencies}
\label{sec:intraclass}
\subsection{Identify model weakness}
\begin{figure}
\centering
\includegraphics[width=0.7\linewidth]{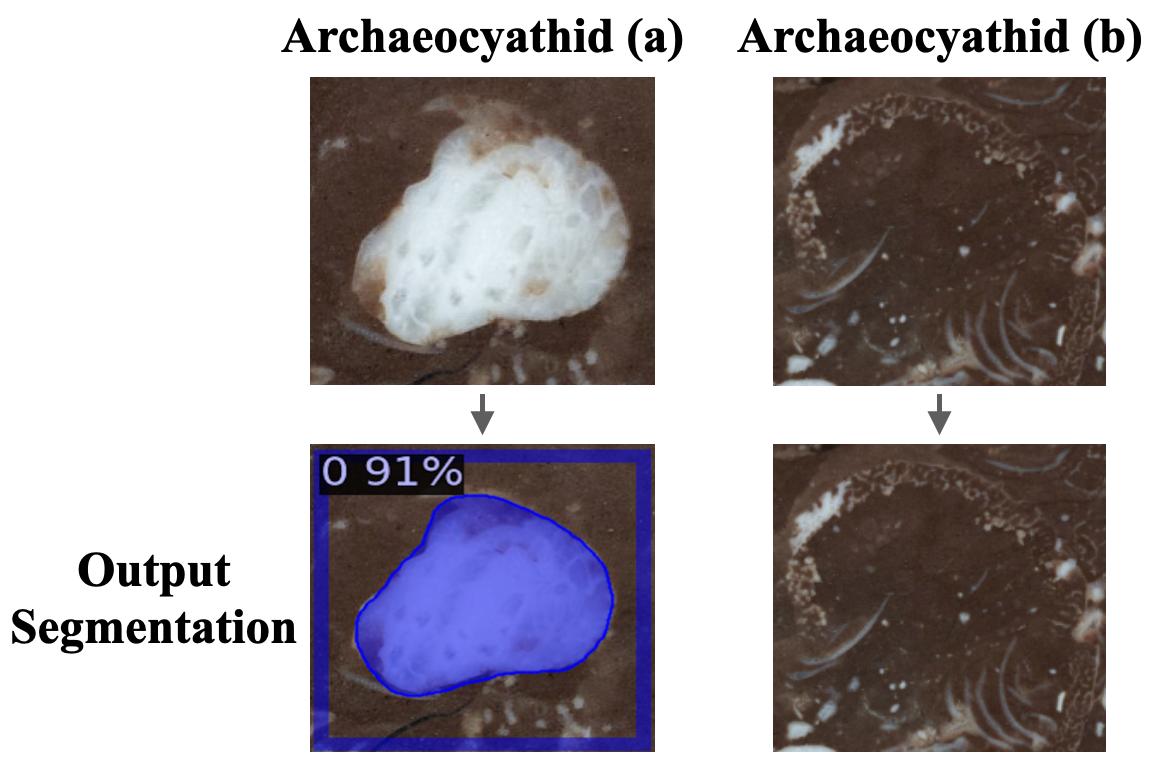}
\caption{\textbf{Inconsistent segmentation of archaeocyathids.} Examples of output archaeocyathids masks with varying segmentation quality. The Mask R-CNN fails to segment the irregular, red mud filled (b) archaeocyathid but produces a complete mask for the regular, (a) archaeocyathid that contains a porous texture.}
\label{fig:texturesegmentations}
\end{figure}

\paragraph{Archaeocyathids can have different textures.}
There exists a fair amount of intraclass variation among archaeocyathids. 
For example, there are recrystallized (white/gray) and red mud filled (red/brown) archaeocyathids, irregular (long) and regular (round) archaeocyathids, and more. 
We denote the two primary textures as \textbf{(a)} and \textbf{(b)} (Fig.~\ref{fig:fewfossils}).
The Mask R-CNN segments \textbf{(a)} (archaeocyathids with porous textures; mean precision $=0.84$; mean recall $=0.67$) better than it segments \textbf{(b)} (archaeocyathids filled with red mud; mean precision $=0.56$; mean recall $=0.25$) (Fig.~\ref{fig:texturesegmentations}).

\begin{figure}[hbt]
\centering
\includegraphics[width=0.6\linewidth]{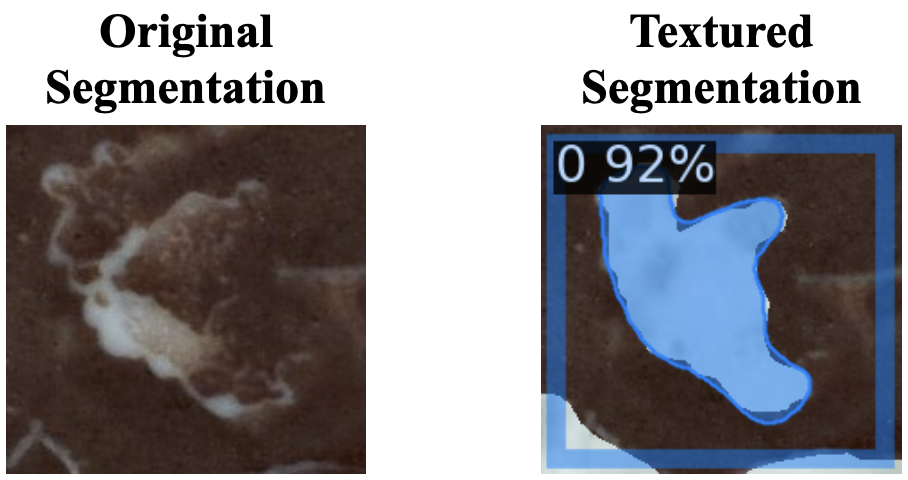}
\caption{\textbf{Example of improvement due to texture replacement.} An example of a red mud filled archaeocyathid that the model misses in the unperturbed image (left) and perfectly segments when replaced with a porous texture (right). A similar trend occurs for other red mud filled archaeocyathids.}
\label{fig:pittedtextureinference}
\end{figure}

\vspace{-1em}
\paragraph{Test effect of optimal texture.}
To test the effect of the porous texture on the segmentation quality, we stitch copies of a square crop of the texture from one such porous-textured archaeocyathid to form a continuous textured image of the same size as our images.
We then substitute the texture in for all the archaeocyathids in the training images (Fig.~\ref{fig:goodtextures}) and run inference on the modified images.
The quality of the masks improves (mean IoU increases from $0.63 \pm 0.29$ to $0.66 \pm 0.31$) though to a lesser extent than the predicted masks for the inpainted non-archaeocyathids.
Further analysis shows that there is an increase both in instance-level precision (mean increases from $0.78 \pm 0.22$ to $0.80 \pm 0.17$) and recall (mean increases from $0.77 \pm 0.25$ to $0.87 \pm 0.23$).
Qualitatively, we find that the segmentation of previously poorly-segmented archaeocyathids improves (Fig.~\ref{fig:pittedtextureinference}).

\subsection{Mitigate model weakness}
\paragraph{Mapping archaeocyathids to the optimal texture.}
Since the modified images seem to solicit an improved segmentation, we apply the model-editing method to map poorly-performing archaeocyathids to the porous texture equivalent. 
Unlike our previous extension of the model-editing method to enforce a binary supercategorization, we use the method to strengthen the characteristics of the archaeocyathid class (i.e. a similar reason as the original work~\cite{editingclassifier}).
We edit with each of the 6 training images individually and produce two additional models edited with image $D$ to test the effect of replacing different amounts of archaeocyathid pixels with the porous texture.

\paragraph{Results}
Although mapping all the archaeocyathids to the porous texture at once sometimes improves the segmentation of the edit image itself, it generally produces masks with lower IoUs and does not improve the precision or recall for the unseen images (\cref{tab:archresults}).
Furthermore, mapping fewer archaeocyathid pixels to the porous texture does not significantly change the performance from the original model (\cref{tab:onearchresults}).
Thus, mapping the archaeocyathids to the porous texture does not seem to be an effective approach.
This trend likely occurs because the visual contrast between archaeocyathids and the porous texture is less vivid than that between non-archaeocyathids and red mud (Sec. \ref{sec:interclass}).



\setlength{\tabcolsep}{5pt}
\begin{table}[t!]
  \centering
  \begin{tabular}{c|c|c|c}
    \toprule
    Image & Precision & Recall & IoU\\
    \midrule
    None &
    $\mathbf{0.86 \pm 0.17}$ &
    $\mathbf{0.75 \pm 0.26}$ & 
    $\mathbf{0.63 \pm 0.28}$\\
    \midrule
    
    A & 
    $0.85 \pm 0.16$ &
    $0.70 \pm 0.30$ & 
    $0.56 \pm 0.32$ \\
    
    B & 
    $0.84 \pm 0.16$ &
    $0.70 \pm 0.29$ &
    $0.55 \pm 0.32$ \\
    
    C & 
    $0.85 \pm 0.18$ &
    $0.73 \pm 0.29$ &
    $0.59 \pm 0.31$ \\
    
    D & 
    $0.84 \pm 0.15$ &
    $0.72 \pm 0.28$ &
    $0.55 \pm 0.32$ \\
    
    E & 
    $0.80 \pm 0.20$ &
    $0.69 \pm 0.32$ &
    $0.52 \pm 0.34$ \\
    
    F & 
    $0.85 \pm 0.15$ &
    $0.70 \pm 0.30$ &
    $0.57 \pm 0.32$ \\
    
    \bottomrule
  \end{tabular}
  \caption{\textbf{Mapping all archaeocyathids to porous texture.}
  Metrics computed on 215 archaeocyathids from 4 images (mean and standard deviation reported) when editing with each of the training images.
  The top row shows the original model's performance (no edits). 
  None of the edits produce an improvement over the original model.
  }
  \label{tab:archresults}
\end{table}

\setlength{\tabcolsep}{6pt}
\begin{table}[t!]
  \centering
  \begin{tabular}{c|c|c|c}
    \toprule
        \% Pixels & Precision & Recall & IoU\\
    \midrule
    None &
    $\mathbf{0.86 \pm 0.17}$ &
    $\mathbf{0.75 \pm 0.26}$ & 
    $\mathbf{0.63 \pm 0.28}$\\
    \midrule
    6 &
    $0.86 \pm 0.16$ &
    $0.75 \pm 0.26$ & 
    $0.62 \pm 0.29$\\
    
    49 & 
    $0.85 \pm 0.16$ &
    $0.74 \pm 0.26$ &
    $0.61 \pm 0.29$ \\
    
    100 & 
    $0.84 \pm 0.15$ &
    $0.72 \pm 0.28$ &
    $0.55 \pm 0.32$ \\
    
    \bottomrule
  \end{tabular}
  \caption{\textbf{Mapping different amounts of archaeocyathid pixels to porous texture.} 
  Metrics computed on 215 archaeocyathids from 4 images (mean and standard deviation reported).
  ``\% Pixels" indicates the percent of archaeocyathid pixels in image $D$ that were replaced with the porous texture.
  The first row shows the original model's performance (no edits).
  The second row is when pixels for one, \textbf{(b)} archaeocyathid are replaced.
  The third row is when pixels for 18 \textbf{(b)} archaeocyathids (roughly $50\%$ of archaeocyathid pixels) are replaced.
  The last row is when all archaeocyathid pixels are replaced.
  The performance for 6\% of replaced pixels is nearly identical to that under the original model; none of the edits show an improvement over the original model. 
  }
  \label{tab:onearchresults}
\end{table}

\section{Combinations of Mappings}
\label{sec:combinations}

\subsection{Simultaneous Mapping}
When we run inference on training images where both non-archaeocyathids are inpainted with red mud and archaeocyathids are replaced with the porous texture (Fig.~\ref{fig:inpaintsimultaneous}), the segmentation improves (mean IoU improves from $0.63 \pm 0.29$ to $0.71 \pm 0.30$, mean precision improves from $0.79 \pm 0.22$ to $0.85 \pm 0.17$, and mean recall improves from $0.77 \pm 0.25$ to $0.88 \pm 0.21$).

However, when we edit ($20k$ steps; $lr = 10^{-4}$) with these images, the edited model produces lower quality masks for both the edit image and the unseen images (\cref{tab:simultaneousresults}).
Thus, this mapping is not an effective approach.

\setlength{\tabcolsep}{2pt}
\begin{table}[h!]
  \centering
  \begin{tabular}{c|c|c|c}
    \toprule
    Image & Precision & Recall & IoU\\
    \midrule
    None &
    $\mathbf{0.86 \pm 0.17}$ &
    $\mathbf{0.75 \pm 0.26}$ & 
    $\mathbf{0.63 \pm 0.28}$\\
    \midrule
    
    A & 
    $0.82 \pm 0.19$ &
    $0.58 \pm 0.29$ & 
    $0.41 \pm 0.30$ \\
    
    B & 
    $0.82 \pm 0.18$ &
    $0.58 \pm 0.29$ &
    $0.41 \pm 0.31$ \\
    
    C & 
    $0.81 \pm 0.22$ &
    $0.57 \pm 0.32$ &
    $0.44 \pm 0.32$ \\
    
    D & 
    $0.82 \pm 0.20$ &
    $0.61 \pm 0.29$ &
    $0.45 \pm 0.32$ \\
    
    E & 
    $0.76 \pm 0.26$ &
    $0.67 \pm 0.30$ &
    $0.43 \pm 0.33$ \\
    
    F & 
    $0.79 \pm 0.24$ &
    $0.66 \pm 0.29$ &
    $0.48 \pm 0.33$ \\
    
    \bottomrule
  \end{tabular}
  \caption{\textbf{Mapping simultaneously.} 
  Mean and standard deviation are computed on 215 archaeocyathids from 4 images when editing with each of the training images.
  The top row indicates the original model's performance on the test images.
  None of the edits produce an improvement over the original model.}
  \label{tab:simultaneousresults}
\end{table}


\subsection{Sequential Mapping}
In addition to simultaneously mapping to both perturbations, we try mapping the non-archaeocyathids to red mud and then mapping the archaeocyathids to the porous texture.
This procedure is identical to editing on multiple images with the non-archaeocyathids to red mud perturbation (Sec.~\ref{sec:nonarchextension}) except we edit ($20k$ steps; $lr = 10^{-4}$) to an image with only the non-archaeocyathids inpainted followed by an image with only the archaeocyathids inpainted (and vice versa).
We perform this mapping with image $C$ since it produces the most improvement in the non-archaeocyathids to red mud mapping (Tab. \ref{tab:nonarchresults}).
We find that the performance of each sequentially edited model corresponds to the performance of the last mapping in isolation. 
For example, the model edited with the non-archaeocyathid mapping followed by the archaeocyathid mapping produces masks of similar quality to the model edited with the archaeocyathid mapping alone (\cref{tab:alternate}).
This trend seems reasonable given the results from the earlier multi-image edit experiments.

\setlength{\tabcolsep}{6pt}
\begin{table}
  \centering
  \begin{tabular}{c|c|c|c}
    \toprule
        Order & Precision & Recall & IoU\\
    \midrule
    \textit{Ar} &
    $0.85 \pm 0.18$ &
    $0.73 \pm 0.29$ &
    $0.59 \pm 0.31$ \\
    
    \textit{No} &
    $0.91 \pm 0.12$ &
    $0.63 \pm 0.26$ &
    $0.56 \pm 0.27$ \\
    \midrule
    
    \textit{Ar, No} &
    $0.91 \pm 0.12$ &
    $0.62 \pm 0.27$ & 
    $0.56 \pm 0.28$\\
    
    \textit{No, Ar} &
    $0.85 \pm 0.16$ &
    $0.73 \pm 0.28$ & 
    $0.59 \pm 0.31$ \\
    \bottomrule
  \end{tabular}
  \caption{\textbf{Mapping sequentially.} 
  Metrics computed on 215 archaeocyathids from 4 images (mean and standard deviation reported); all edits were done using only image $C$. 
  \textit{Ar, No} represents mapping archaeocyathids to porous texture followed by mapping non-archaeocyathids to red mud; \textit{No, Ar} represents the reverse sequence.
  \textit{Ar} and \textit{No} results are from~\cref{tab:archresults,tab:nonarchresults}.
  The result of each sequential mapping is similar to the result when editing with the last mapping only (i.e. \textit{Ar, No} is similar to \textit{No}).
  }
  \label{tab:alternate}
\end{table}

\section{Conclusion}
\label{sec:conclusion}
In this work, we focus on improving a fossil segmentation model first by identifying its model weaknesses via image perturbations and second by mitigating those weaknesses using model editing.
Specifically, we study a Mask R-CNN trained on a small, fine-grain rock sample dataset to segment instances of archaeocyathid fossils.

First, we show how inpainting and texture synthesis can identify model weaknesses such as interclass confusion (e.g.\ our model confused a different type of fossil for archaeocyathids) and intraclass inconsistencies (e.g.\ performance varied for archaeocyathids with different textures).
Second, we extend a model-editing technique~\cite{editingclassifier} for image classification to image segmentation and show how to best apply it to mitigate identified model weaknesses.
We show that one edit image is sufficient, that mapping all relevant pixels is more effective than mapping fewer pixels, and that sequentially performing edits typically yields the same performance as the last edit alone.

We also demonstrate that model editing may not work in challenging circumstances when the visual appearance of an object is very similar to that of another type of object in the edit image.
Lastly, we find that while model-editing can negatively impact IoU and recall, it can improve precision when designed to mitigate interclass confusion (e.g. treating non-archaeocyathid pixels as red mud).

Although our work focuses on improving a fossil segmentation model, our methodology may be useful for tackling similar problems that involve training a segmentation model on a small, fine-grained dataset.
Further research could investigate what properties of the images cause one training image to be more effective than another and work towards mitigating the negative impacts on IoU and recall. 

\vspace{-1em}
\paragraph{Limitations.}
Given that our work focuses on editing a Mask R-CNN trained on a few images from one rock sample, the main limitation is that our findings may not generalize well to other segmentation models trained on small, fine-grain datasets.
Our goal is to present a novel combination of techniques for investigating and improving the performance of segmentation models with a limited amount of labeled data, and we ran extensive experiments to substantiate our decisions.
Thus, work in novel domains should validate our findings on their own models and datasets.\blfootnote{Acknowledgements: We are grateful for support from the Princeton SEAS IW funding (IP), the Princeton SEAS Project X Fund (RF), and Open Philanthropy (RF). We thank Devon Ulrich and Sunnie S. Y. Kim for helpful discussions.}

{\small
\bibliographystyle{ieee_fullname}
\bibliography{egbib}

\begin{thebibliography}{10}\itemsep=-1pt

\bibitem{googleblog}
{Fluid Annotation: An Exploratory Machine Learning–Powered Interface for
  Faster Image Annotation}.
\newblock
  \url{https://ai.googleblog.com/2018/10/fluid-annotation-exploratory-machine.html}.
\newblock Accessed: 2022-06-24.

\bibitem{beutelfair}
Alex Beutel, Jilin Chen, Zhe Zhao, and Ed~H. Chi.
\newblock Data decisions and theoretical implications when adversarially
  learning fair representations.
\newblock {\em arXiv}, 2017.

\bibitem{botany}
Laura Brenskelle, Rob~P Guralnick, Michael Denslow, and Brian~J Stucky.
\newblock {Maximizing human effort for analyzing scientific images: A case
  study using digitized herbarium sheets}.
\newblock {\em Applications in plant sciences}, 8(6):e11370--e11370, July 2020.

\bibitem{cocoannotator}
Justin Brooks.
\newblock {COCO Annotator}.
\newblock \url{https://github.com/jsbroks/coco-annotator/}, 2019.

\bibitem{devries2017improved}
Terrance DeVries and Graham~W Taylor.
\newblock Improved regularization of convolutional neural networks with cutout.
\newblock {\em arXiv}, 2017.

\bibitem{fong17interpretable}
Ruth Fong and Andrea Vedaldi.
\newblock Interpretable explanations of black boxes by meaningful perturbation.
\newblock In {\em {ICCV}}, 2017.

\bibitem{fong2019occlusions}
Ruth Fong and Andrea Vedaldi.
\newblock Occlusions for effective data augmentation in image classification.
\newblock In {\em {ICCV} Workshop}, 2019.

\bibitem{connectomicsex1}
Jan Funke, Fabian Tschopp, William Grisaitis, Arlo Sheridan, Chandan Singh,
  Stephan Saalfeld, and Srinivas~C. Turaga.
\newblock Large scale image segmentation with structured loss based deep
  learning for connectome reconstruction.
\newblock {\em {PAMI}}, 41(7):1669--1680, 2019.

\bibitem{gatys2015texture}
Leon Gatys, Alexander~S Ecker, and Matthias Bethge.
\newblock Texture synthesis using convolutional neural networks.
\newblock {\em {NeurIPS}}, 2015.

\bibitem{geirhos2018imagenettrained}
Robert Geirhos, Patricia Rubisch, Claudio Michaelis, Matthias Bethge, Felix~A.
  Wichmann, and Wieland Brendel.
\newblock Imagenet-trained {CNN}s are biased towards texture; increasing shape
  bias improves accuracy and robustness.
\newblock In {\em {ICLR}}, 2019.

\bibitem{maskrcnn}
Kaiming He, Georgia Gkioxari, Piotr Dollar, and Ross Girshick.
\newblock {Mask R-CNN}.
\newblock In {\em {ICCV}}, 2017.

\bibitem{ilharco2022}
Gabriel Ilharco, Marco~Tulio Ribeiro, Mitchell Wortsman, Suchin Gururangan,
  Ludwig Schmidt, Hannaneh Hajishirzi, and Ali Farhadi.
\newblock Editing models with task arithmetic, 2022.

\bibitem{texturesynthesis}
Xuejing Lei, Ganning Zhao, and C.~C.~Jay Kuo.
\newblock Nites: A non-parametric interpretable texture synthesis method.
\newblock {\em arXiv}, 2020.

\bibitem{deepimageprior}
Victor Lempitsky, Andrea Vedaldi, and Dmitry Ulyanov.
\newblock Deep image prior.
\newblock In {\em {CVPR}}, 2018.

\bibitem{fpn}
Tsung-Yi Lin, Piotr Dollár, Ross Girshick, Kaiming He, Bharath Hariharan, and
  Serge Belongie.
\newblock {Feature Pyramid Networks for Object Detection}, 2016.

\bibitem{cocodataset}
Tsung-Yi Lin, Michael Maire, Serge Belongie, James Hays, Pietro Perona, Deva
  Ramanan, Piotr Dollar, and Larry Zitnick.
\newblock {Microsoft COCO: Common Objects in Context}.
\newblock In {\em {ECCV}}, 2014.

\bibitem{connectomicsex2}
Drew Linsley, Junkyung Kim, David~M. Berson, and Thomas Serre.
\newblock Robust neural circuit reconstruction from serial electron microscopy
  with convolutional recurrent networks.
\newblock {\em arXiv}, 2018.

\bibitem{laftr}
David Madras, Elliot Creager, Toniann Pitassi, and Richard Zemel.
\newblock Learning adversarially fair and transferable representations.
\newblock In {\em {ICML}}, 2018.

\bibitem{manzuk}
Ryan~A. Manzuk, Adam~C Maloof, Jaap~A Kaandorp, and Mark Webster.
\newblock {{Branching archaeocyathids as ecosystem engineers during the
  Cambrian radiation}}.
\newblock {\em Geobiology}, pages 1--20, 2022.

\bibitem{connectomicsex3}
Brian Matejek, Daniel Haehn, Haidong Zhu, Donglai Wei, Toufiq Parag, and
  Hanspeter Pfister.
\newblock Biologically-constrained graphs for global connectomics
  reconstruction.
\newblock In {\em {CVPR}}, 2019.

\bibitem{imagingrocks}
Akshay Mehra, Bolton Howes, Ryan Manzuk, Alex Spatzier, Bradley~M Samuels, and
  Adam~C Maloof.
\newblock A novel technique for producing three-dimensional data using serial
  sectioning and semi-automatic image classification.
\newblock {\em Microscopy and Microanalysis}, 28(6):2020--2035, 2022.

\bibitem{cloudina}
Akshay Mehra and Adam Maloof.
\newblock {Multiscale approach reveals that Cloudina aggregates are detritus
  and not in situ reef constructions}.
\newblock {\em {PNAS}}, 115(11):E2526, 2018.

\bibitem{meng2022locating}
Kevin Meng, David Bau, Alex Andonian, and Yonatan Belinkov.
\newblock Locating and editing factual associations in {GPT}.
\newblock {\em Advances in Neural Information Processing Systems}, 36, 2022.

\bibitem{meng2022memit}
Kevin Meng, Arnab Sen~Sharma, Alex Andonian, Yonatan Belinkov, and David Bau.
\newblock Mass editing memory in a transformer.
\newblock {\em arXiv preprint arXiv:2210.07229}, 2022.

\bibitem{mordvintsev2018differentiable}
Alexander Mordvintsev, Nicola Pezzotti, Ludwig Schubert, and Chris Olah.
\newblock Differentiable image parameterizations.
\newblock {\em Distill}, 3(7):e12, 2018.

\bibitem{Petsiuk2018rise}
Vitali Petsiuk, Abir Das, and Kate Saenko.
\newblock Rise: Randomized input sampling for explanation of black-box models.
\newblock In {\em {BMVC}}, 2018.

\bibitem{ribeiro2016should}
Marco~Tulio Ribeiro, Sameer Singh, and Carlos Guestrin.
\newblock ``{W}hy should {I} trust you?'' explaining the predictions of any
  classifier.
\newblock In {\em {KDD}}, 2016.

\bibitem{coralbleaching}
Eugene Rosenberg, Omry Koren, Leah Reshef, Rotem Efrony, and Ilana
  Zilber-Rosenberg.
\newblock {The role of microorganisms in coral health, disease and evolution}.
\newblock {\em Nature Reviews Microbiology}, 5(5):355--362, 2007.

\bibitem{archaeos}
Stephen~M. Rowland and Roland~A. Gangloff.
\newblock Structure and paleoecology of lower cambrian reefs.
\newblock {\em PALAIOS}, 3(2):111--135, 1988.

\bibitem{imagenet}
Olga Russakovsky, Jia Deng, Hao Su, Jonathan Krause, Sanjeev Satheesh, Sean Ma,
  Zhiheng Huang, Andrej Karpathy, Aditya Khosla, Michael Bernstein, Alexander~C
  Berg, and Li Fei-Fei.
\newblock {ImageNet Large Scale Visual Recognition Challenge}.
\newblock {\em {IJCV}}, 2015.

\bibitem{geosciences}
Jéssica~S. Santos, Rodrigo~S. Ferreira, and Viviane~T. Silva.
\newblock Evaluating the classification of images from geoscience papers using
  small data.
\newblock {\em Applied Computing and Geosciences}, 5:100018, 2020.

\bibitem{editingclassifier}
Shibani Santurkar, Dimitris Tsipras, Mahalaxmi Elango, David Bau, Antonio
  Torralba, and Aleksander Madry.
\newblock Editing a classifier by rewriting its prediction rules.
\newblock In {\em {NeurIPS}}, 2021.

\bibitem{hideandseek}
K.K. Singh and Y.J. Lee.
\newblock {Hide-and-Seek: Forcing a Network to be Meticulous for
  Weakly-Supervised Object and Action Localization}.
\newblock In {\em ICCV}, 2017.

\bibitem{sinitsin2020}
Anton Sinitsin, Vsevolod Plokhotnyuk, Dmitry Pyrkin, Sergei Popov, and Artem
  Babenko.
\newblock Editable neural networks.
\newblock In {\em International Conference on Learning Representations}, 2020.

\bibitem{adv_fairness}
Christina Wadsworth, Francesca Vera, and Chris Piech.
\newblock Achieving fairness through adversarial learning: an application to
  recidivism prediction.
\newblock {\em arXiv}, 2018.

\bibitem{medicine}
Shanshan Wang, Cheng Li, Rongpin Wang, Zaiyi Liu, Meiyun Wang, Hongna Tan,
  Yaping Wu, Xinfeng Liu, Hui Sun, Rui Yang, Xin Liu, Jie Chen, Huihui Zhou,
  Ismail {Ben Ayed}, and Hairong Zheng.
\newblock {Annotation-efficient deep learning for automatic medical image
  segmentation}.
\newblock {\em Nature Communications}, 12(1):5915, 2021.

\bibitem{wang2018balanced}
Tianlu Wang, Jieyu Zhao, Mark Yatskar, Kai-Wei Chang, and Vicente Ordonez.
\newblock Balanced datasets are not enough: Estimating and mitigating gender
  bias in deep image representations.
\newblock {\em arXiv}, 2018.

\bibitem{wei2017object}
Yunchao Wei, Jiashi Feng, Xiaodan Liang, Ming-Ming Cheng, Yao Zhao, and
  Shuicheng Yan.
\newblock Object region mining with adversarial erasing: A simple
  classification to semantic segmentation approach.
\newblock In {\em {CVPR}}, 2017.

\bibitem{detectron2}
Yuxin Wu, Alexander Kirillov, Francisco Massa, Wan-Yen Lo, and Ross Girshick.
\newblock Detectron2.
\newblock \url{https://github.com/facebookresearch/detectron2}, 2019.

\bibitem{zeiler2014visualizing}
Matthew~D Zeiler and Rob Fergus.
\newblock Visualizing and understanding convolutional networks.
\newblock In {\em {ECCV}}, 2014.

\bibitem{connectomicsoverview}
Tao Zeng, Bian Wu, and Shuiwang Ji.
\newblock {DeepEM3D: approaching human-level performance on 3D anisotropic EM
  image segmentation}.
\newblock {\em Bioinformatics}, 33(16):2555--2562, aug 2017.

\bibitem{zhang2018mitigating}
Brian~Hu Zhang, Blake Lemoine, and Margaret Mitchell.
\newblock Mitigating unwanted biases with adversarial learning.
\newblock In {\em {AEIS}}, 2018.

\bibitem{occludeuntilmisclassify}
Bolei Zhou, Aditya Khosla, {\`A}gata Lapedriza, Aude Oliva, and Antonio
  Torralba.
\newblock {Object Detectors Emerge in Deep Scene CNNs}.
\newblock {\em {ICLR}}, 2015.

\bibitem{zhu2020}
Chen Zhu, Ankit~Singh Rawat, Manzil Zaheer, Srinadh Bhojanapalli, Daliang Li,
  Felix Yu, and Sanjiv Kumar.
\newblock Modifying memories in transformer models, 2020.

\end{thebibliography}
}

\newpage
\onecolumn
\section*{Appendix}
\label{sec:appendix}
This appendix contains tables for evaluation on the edit images (i.e. each of the 6 training images). We include these evaluations to check if the edited models perform better than the original model on the image used for the corresponding edit. The evaluations on the unseen images are included in the main paper. In general, we see that the performance of each mapping on the edit images loosely corresponds to its performance on the unseen images. 
\begin{table*}[h]
  \centering
  \begin{tabular}{c|c|c|c}
    \toprule
    Image & Precision & Recall & IoU \\
    \midrule
     & Original \hspace*{1.2cm} Edited & Original \hspace*{1.2cm} Edited & Original \hspace*{1.2cm} Edited \\
    \midrule
    A & 
    $0.82 \pm 0.19 \hspace*{0.5cm} \mathbf{0.88 \pm 0.19}$ &
    $\mathbf{0.78 \pm 0.24} \hspace*{0.5cm} 0.63 \pm 0.28$ & 
    $\mathbf{0.66 \pm 0.25} \hspace*{0.5cm} 0.57 \pm 0.29$ \\
    
    B & 
    $0.53 \pm 0.28 \hspace*{0.5cm} \mathbf{0.54 \pm 0.27}$ &
    $\mathbf{0.78 \pm 0.24} \hspace*{0.5cm} 0.67 \pm 0.24$ &
    $\mathbf{0.66 \pm 0.27} \hspace*{0.5cm} 0.61 \pm 0.26$ \\
    
    C & 
    $0.83 \pm 0.20 \hspace*{0.5cm} \mathbf{0.95 \pm 0.09}$ &
    $\mathbf{0.77 \pm 0.24} \hspace*{0.5cm} 0.70 \pm 0.23$ &
    $0.64 \pm 0.30 \hspace*{0.5cm} \mathbf{0.67 \pm 0.24}$ \\
    
    D & 
    $0.79 \pm 0.23 \hspace*{0.5cm} \mathbf{0.91 \pm 0.13}$ &
    $\mathbf{0.78 \pm 0.28} \hspace*{0.5cm} 0.74 \pm 0.23$ &
    $0.64 \pm 0.30 \hspace*{0.5cm} \mathbf{0.66 \pm 0.26}$ \\
    
    E & 
    $0.85 \pm 0.15 \hspace*{0.5cm} \mathbf{0.92 \pm 0.10}$ &
    $\mathbf{0.70 \pm 0.26} \hspace*{0.5cm} 0.63 \pm 0.28$ &
    $0.55 \pm 0.32 \hspace*{0.5cm} \mathbf{0.58 \pm 0.29}$ \\
    
    F & 
    $0.83 \pm 0.16 \hspace*{0.5cm} \mathbf{0.90 \pm 0.10}$ &
    $\mathbf{0.82 \pm 0.22} \hspace*{0.5cm} 0.72 \pm 0.21$ &
    $\mathbf{0.67 \pm 0.26} \hspace*{0.5cm} 0.64 \pm 0.23$ \\
    
    \bottomrule
  \end{tabular}
  \caption{\textbf{Mapping non-archaeocyathids to red mud: Performance on edit image.} This table contains the mean instance-level metrics $\pm$ one standard deviation run on each image used for mapping non-archaeocyathids to red mud. For example, the first row contains the mean instance-level precision, recall, and IoU across the identified archaeocyathids in image $A$ alone. All the precisions increase, and some of the IoU scores increase as well. The recall scores decrease.}
  \label{tab:nonarchresultstuningimage}
\end{table*}

\begin{table*}[h]
  \centering
  \begin{tabular}{c|c|c|c}
    \toprule
    Image & Precision & Recall & IoU \\
    \midrule
     & Original \hspace*{1.2cm} Edited & Original \hspace*{1.2cm} Edited & Original \hspace*{1.2cm} Edited \\
    \midrule
    A & 
    $\mathbf{0.82 \pm 0.19} \hspace*{0.5cm} \mathbf{0.82 \pm 0.18}$ &
    $0.78 \pm 0.24 \hspace*{0.5cm} \mathbf{0.81 \pm 0.22}$ & 
    $0.66 \pm 0.25 \hspace*{0.5cm} \mathbf{0.67 \pm 0.26}$ \\
    
    B & 
    $\mathbf{0.53 \pm 0.28} \hspace*{0.5cm} \mathbf{0.53 \pm 0.25}$ &
    $\mathbf{0.78 \pm 0.24} \hspace*{0.5cm} 0.77 \pm 0.24$ &
    $\mathbf{0.66 \pm 0.27} \hspace*{0.5cm} 0.62 \pm 0.29$ \\
    
    C & 
    $0.83 \pm 0.20 \hspace*{0.5cm} \mathbf{0.84 \pm 0.15}$ &
    $\mathbf{0.77 \pm 0.24} \hspace*{0.5cm} 0.75 \pm 0.30$ &
    $\mathbf{0.64 \pm 0.30} \hspace*{0.5cm} 0.60 \pm 0.34$ \\
    
    D & 
    $0.79 \pm 0.23 \hspace*{0.5cm} \mathbf{0.85 \pm 0.16}$ &
    $0.78 \pm 0.28 \hspace*{0.5cm} \mathbf{0.82 \pm 0.26}$ &
    $0.64 \pm 0.30 \hspace*{0.5cm} \mathbf{0.66 \pm 0.28}$ \\
    
    E & 
    $\mathbf{0.85 \pm 0.15} \hspace*{0.5cm} 0.80 \pm 0.16$ &
    $0.70 \pm 0.26 \hspace*{0.5cm} \mathbf{0.73 \pm 0.30}$ &
    $\mathbf{0.55 \pm 0.32} \hspace*{0.5cm} 0.48 \pm 0.35$ \\
    
    F & 
    $\mathbf{0.83 \pm 0.16} \hspace*{0.5cm} 0.82 \pm 0.14$ &
    $\mathbf{0.82 \pm 0.22} \hspace*{0.5cm} 0.81 \pm 0.26$ &
    $\mathbf{0.67 \pm 0.26} \hspace*{0.5cm} 0.65 \pm 0.28$ \\
    
    \bottomrule
  \end{tabular}
  \caption{\textbf{Mapping all archaeocyathids to pitted texture: Performance on edit image.} This table contains the mean instance-level metrics $\pm$ one standard deviation run on each image used for mapping archaeocyathids to the pitted texture. For example, the first row contains the mean instance-level precision, recall, and IoU across the identified archaeocyathids in image $A$ alone. There does not appear to be a clear trend in any of the metrics. For example, image $D$ produces an improvement across all metrics, while image $F$ does not.}
  \label{tab:archresultstuningimage}
\end{table*}

\begin{table*}[h]
  \centering
  \begin{tabular}{c|c|c|c}
    \toprule
    Image & Precision & Recall & IoU \\
    \midrule
     & Original \hspace*{1.2cm} Edited & Original \hspace*{1.2cm} Edited & Original \hspace*{1.2cm} Edited \\
    \midrule
    A & 
    $\mathbf{0.82 \pm 0.19} \hspace*{0.5cm} 0.80 \pm 0.21$ &
    $\mathbf{0.78 \pm 0.24} \hspace*{0.5cm} 0.60 \pm 0.32$ & 
    $\mathbf{0.66 \pm 0.25} \hspace*{0.5cm} 0.44 \pm 0.35$ \\
    
    B & 
    $\mathbf{0.53 \pm 0.28} \hspace*{0.5cm} 0.51 \pm 0.32$ &
    $\mathbf{0.78 \pm 0.24} \hspace*{0.5cm} 0.64 \pm 0.28$ &
    $\mathbf{0.66 \pm 0.27} \hspace*{0.5cm} 0.51 \pm 0.32$ \\
    
    C & 
    $\mathbf{0.83 \pm 0.20} \hspace*{0.5cm} \mathbf{0.83 \pm 0.16}$ &
    $\mathbf{0.77 \pm 0.24} \hspace*{0.5cm} 0.64 \pm 0.32$ &
    $\mathbf{0.64 \pm 0.30} \hspace*{0.5cm} 0.44 \pm 0.36$ \\
    
    D & 
    $\mathbf{0.79 \pm 0.23} \hspace*{0.5cm} 0.80 \pm 0.21$ &
    $\mathbf{0.78 \pm 0.28} \hspace*{0.5cm} 0.77 \pm 0.24$ &
    $\mathbf{0.64 \pm 0.30} \hspace*{0.5cm} 0.53 \pm 0.36$ \\
    
    E & 
    $\mathbf{0.85 \pm 0.15} \hspace*{0.5cm} \mathbf{0.85 \pm 0.14}$ &
    $\mathbf{0.70 \pm 0.26} \hspace*{0.5cm} 0.60 \pm 0.32$ &
    $\mathbf{0.55 \pm 0.32} \hspace*{0.5cm} 0.46 \pm 0.34$ \\
    
    F & 
    $0.83 \pm 0.16 \hspace*{0.5cm} \mathbf{0.87 \pm 0.11}$ &
    $\mathbf{0.82 \pm 0.22} \hspace*{0.5cm} 0.74 \pm 0.24$ &
    $\mathbf{0.67 \pm 0.26} \hspace*{0.5cm} 0.64 \pm 0.26$ \\
    
    \bottomrule
  \end{tabular}
  \caption{\textbf{Mapping simultaneously: Performance on edit image.} This table contains the mean instance-level metrics $\pm$ one standard deviation run on each image used for simultaneously mapping non-archaeocyathids to red mud and archaeocyathids to the pitted texture. For example, the first row contains the mean instance-level precision, recall, and IoU across the identified archaeocyathids in image $A$ alone. Only image $F$ produces an increase in precision. None of the edited models have higher recall or IoU scores than the original model.}
  \label{tab:simultaneousresultstuningimage}
\end{table*}

\end{document}